# Detecting Anemia from Retinal Fundus Images

Akinori Mitani, MD, PhD[1]
Yun Liu, PhD[1]
Abigail Huang, MD[1]
Greg S. Corrado, PhD[1]
Lily Peng, MD, PhD[1]
Dale R. Webster, PhD[1]
Naama Hammel, MD[1]
Avinash V. Varadarajan, MS[1]

[1] Google AI Healthcare, Google, Mountain View, CA, USA

Corresponding author:
Akinori Mitani, MD, PhD
Google AI Healthcare
1600 Amphitheatre Parkway
Mountain View, CA 94043
amitani@google.com

## Abstract

Despite its high prevalence, anemia is often undetected due to the invasiveness and cost of screening and diagnostic tests. Though some non-invasive approaches have been developed, they are less accurate than invasive methods, resulting in an unmet need for more accurate non-invasive methods. Here, we show that deep learning-based algorithms can detect anemia and quantify several related blood measurements using retinal fundus images both in isolation and in combination with basic metadata such as patient demographics. On a validation dataset of 11,388 patients from the UK Biobank, our algorithms achieved a mean absolute error of 0.63 g/dL (95% confidence interval (CI) 0.62-0.64) in quantifying hemoglobin concentration and an area under receiver operating characteristic curve (AUC) of 0.88 (95% CI 0.86-0.89) in detecting anemia. This work shows the potential of automated non-invasive anemia screening based on fundus images, particularly in diabetic patients, who may have regular retinal imaging and are at increased risk of further morbidity and mortality from anemia.

## Introduction

Anemia is a public health problem affecting an estimated 1.62 billion people[1]. In 2011, 29% of non-pregnant women worldwide were affected by anemia[2]. As a major contributor to global burden of disease, anemia has far-reaching consequences for work and productivity and quality of life[1,3,4]. As anemia is usually correctable[5], timely detection and intervention are key. The most reliable indicator of anemia is hemoglobin concentration (Hb)[1], which is traditionally measured using a venous or capillary blood sample. However, these procedures are invasive and painful, can cause infection of patients and healthcare workers, and generate biohazardous waste[6]. Thus, there is a clear need for a non-invasive procedure.

Several non-invasive methods of estimating Hb are now available. Traditionally, subjective assessment of pallor of the conjunctiva, nail beds, tongue, and palms have been used as clinical signs indicating the presence of severe anemia, with a wide range of estimated sensitivities and specificities[7,8]. Recently, it was reported that Hb can be estimated with high accuracy using automated algorithms from the color of fingernail beds in digital photographs taken by smartphones[9]. However, the algorithms were based on manually selected regions of interest on the fingernails, and their robustness and real-world performance remains to be assessed. Another two methods, occlusion spectroscopy and pulse-co-oximetry, use spectrophotometric sensors that non-invasively assess Hb by measuring light transmission through the tissue[10,11]. These non-invasive methods are less accurate than the gold standard of venous blood laboratory analysis[12] and represent trade-offs between invasiveness, time, cost, and accuracy[9].

Interestingly, anemia of sufficient severity has been known to manifest characteristic signs in the fundus of the eye[13], and 20% of anemia patients have reported to develop extravascular lesions, with severity of anemia related to venous tortuosity[14]. Retinopathy is observed in 28.3% of patients with anemia and/or thrombocytopenia, and low Hb level was associated with the presence of retinopathy[14]. However, the low prevalence of retinopathy among anemic patients limits its potential sensitivity as a stand-alone diagnostic feature, and fundus photographs have not been used to either detect anemia or quantify more precise Hb levels.

In this work, we explore the hypothesis that Hb can be quantified using noninvasive fundus photographs and deep learning (DL). DL has been previously shown to be highly effective in extracting information from images[15]. In ophthalmology, DL algorithms can detect eye conditions such as diabetic retinopathy, age-related macular degeneration, and glaucoma with accuracy comparable to human experts[16–21]. Additionally, some previously-unknown information can be extracted from fundus images, such as refractive error[22], age, sex, and cardiovascular risk[23]. Extending this, we show that DL can be leveraged to quantify Hb and detect anemia.

## Results

This study was conducted using data from UK Biobank[24]. In total, 114,205 fundus images from 57,163 patients were included in this study. In the validation set, the median age of the patients was 57.9 (50.0-63.8, interquartile range) years, 54.9% were female, and 91.1% were White (Table 1). Among all the patients, 3.7% had anemia, with Hb ranging from 6.4 [g/dL] to 19.6 [g/dL]. Additional patient demographics are summarized in Table 1.

Results of many blood tests including Hb are correlated with patients' metadata such as demographics. Some of these metadata, such as age and gender have previously been shown to be predictable using fundus images[23]. Therefore to later ensure that our predictors were not solely predicting Hb measurements via age and gender, we first developed baseline linear regression models using these metadata (metadata-only model). Next, we developed deep-learning based models using fundus images (fundus-only model). Lastly, we hypothesized that a model utilizing both types of input data may be even more accurate and thus developed combined models that use both metadata and fundus images (combined model).

First, we compared the performance of metadata-only, fundus-only, and combined models trained to predict Hb, hematocrit (HCT), and red blood cell count (RBC), which are correlated with each other and related to anemia (Fig. 1). The mean absolute error (MAE) for predicting Hb by the metadata-only model was 0.73 (95% confidence interval (CI): 0.72-0.74) g/dL. MAE was 0.67 (95% CI: 0.66-0.68) g/dL for the fundus-only model and 0.63 (95% CI: 0.62-0.64) g/dL for the combined model. The performance of predicting HCT and RBC follow a similar trend across the three models. Thus, the combined model predicted Hb, HCT, and RBC more accurately than either fundus-only or metadata-only models, indicating that both metadata and fundus images are important for the accurate prediction.

Additionally, we investigated whether errors made by the model were patient-specific. In the test set, 342 patients had two visits with both fundus images and a Hb measurement. We applied the combined model to the two visits and found that the residual error was correlated between multiple visits over time by the same patient (Fig. 2, Pearson's correlation coefficient r = 0.38 (95% CI: 0.18-0.65)).

Next, we examined if fundus images can be used to predict anemia by developing a DL based classification model to directly predict whether a patient is anemic. Using World Health Organization's (WHO) Hb cutoffs for anemia, we trained the models for three binary classification tasks (anemia, moderate anemia, approximate anemia, see Methods). For all of them, the combined model and the fundus-only model performed better than metadata-only model (Fig. 3, Table 2). The area-under-receiver-operating-characteristic-curve (AUC) for detecting anemia was 0.73 (95% CI: 0.72-0.75) for the metadata-only model, 0.87 (95% CI: 0.85-0.89) for the fundus-only model, and 0.88 (95% CI: 0.86-0.89) for the combined model (Fig. 3a). AUC for detecting moderate anemia was 0.79 (95% CI: 0.76-0.82) for the metadata-only model, 0.95 (95% CI: 0.93-0.97) for the fundus-only model, and 0.95 (95% CI: 0.93-0.97) for the combined model (Fig. 3b). The sensitivity to detect moderate anemia was 57.3% (95% CI: 47.6-70.5) for the metadata-only model, 91.2% (95% CI: 87.7-94.4) for the fundus-only model, and 92.3% (95% CI: 88.1-95.9) for the combined model, when the specificity was 80% (Table 2). These results show that both the fundus-only model and the combined model successfully extracted information about anemia from fundus images, and it supports the hypothesis that DL model can help detect anemia using fundus images.

We further investigated the importance of different anatomic features to the prediction by "ablating" different aspects of the images during both model training and validation (Fig. 4 and 5). Most notably, when the top and bottom parts of the images were masked, the performance started to decline only after about 80% of each image was masked (Fig. 4a, b). Masking the horizontal stripes through the middle of the images decreased the performance after about 20% of each image was masked (Fig. 4d, e). When either the circular central core or the outer rim of the image was masked, the performance started to decline after about 40% was masked (Fig. 4g, h, j, k). The biggest drop in AUC when 10% of the image was masked occurred while masking the central horizontal stripe: an AUC drop of about 3%. We also examined the effect of Gaussian blurs (Fig. 5), and found that applying a Gaussian blur with $\sigma = 8$ pixels decreased the AUC for predicting moderate anemia from 0.92 to 0.83 (Fig. 5c).

Lastly, we examined if other components of the complete blood count (CBC) could be predicted from fundus images. Anemia is diagnosed based on blood Hb measurements, which is often measured as a part of the CBC. In addition to Hb, HCT, and RBC, other measurements such as the mean corpuscular volume (MCV) also help to diagnose anemia and identify the subtype. Since the different components of CBC are measured on separate scales, we compared the model's performance across tests with the $R^2$ coefficient of determination. The combined model was not able to predict MCV with high accuracy, with a low $R^2$ of 0.12. On the other hand, the model predicted the three anemia-related measurements (Hb, HCT, and RBC) most accurately, $R^2$ values of 0.52, 0.49, and 0.36, respectively (Table S1).

## Discussion

This study shows that a DL-based approach leveraging retinal fundus images and metadata can both detect anemia and quantify Hb measurements, potentially enabling automated anemia screening using fundus images.

To help put the accuracy of our and other non-invasive anemia detection methods in context, it can be useful to consider the variability of the ground truth itself[25,26]. As consistent with other studies, the ground truth in this study was Hb measured by laboratory hematology analyzers[27]. The standard deviation of the difference between hematology analyzers and the hemoglobincyanide method (HiCN, the gold standard for Hb measurement for research) was 0.18 g/dL[28], equivalent to MAE of 0.14 g/dL. Thus a portion of our model's MAE (0.63 g/dL) may be attributable to variability in the laboratory Hb measurement. The accuracy of our approach is also comparable to invasive (pooled standard deviation of difference was 0.64 g/dL[29], equivalent to MAE of 0.51 g/dL) and non-invasive point-of-care devices (pooled standard deviation was 1.4 g/dL[29,30], equivalent to MAE of 1.1 g/dL), and non-invasive smartphone-based application (95% limits of agreement was 2.4 g/dL[9], equivalent to MAE of 0.96 g/dL) (Table 3).

In addition, because fundus photographs are routinely captured as part of teleretinal screening for diabetic retinopathy[31–33], the ability to predict Hb from these photographs may provide a convenient way for anemia screening in diabetics with minimal additional cost. The clinical motivation and utility of anemia screening in diabetics is many-fold. First, diabetics can develop chronic kidney disease, which leads to renal anemia[34]. Second, in patients with diabetic kidney disease, renal anemia often develops earlier than other forms of nephropathy[35–38]. Even in the absence of nephropathy, Hb in diabetic patients tends to decrease over time[34,39]. Additionally, the correction of anemia improves the quality of life and may reduce diabetic complications[38]. Finally, from an ophthalmic standpoint, anemia is an independent risk factor for developing high-risk proliferative diabetic retinopathy[40]. Thus, regular screening for anemia is recommended for diabetic patients[40–42].

Beyond the initial diagnosis, subsequent Hb measurements are used to track progression of anemia and response to treatment. In addition, even without a diagnosis of anemia, rapid decrease in Hb may indicate existence or onset of an underlying disease. Thus, in addition to the Hb level at one visit, the difference in Hb across visits is clinically important. Interestingly, our prediction error between multiple visits of the same patient was correlated, showing that some component of the error is patient-specific. When comparing the differences in predictions across multiple visits, the patient-specific components cancel out. Thus, our approach may provide additional value in monitoring the trend in Hb. This premise will be further supported by future work, such as assessing the time delay between true Hb changes and changes in Hb predicted by the algorithm using fundus images: whether the prediction of the model reflects instantaneous Hb or the average Hb over a certain time window. Understanding how the algorithm works would help answer this question. If the algorithm is quantifying the degree of pallor in the fundus, we would expect minimal to no time delays in the order of minutes or hours. On the other hand, if the algorithm is examining features in microvasculature that develop over time, the time delay may be weeks or months. Investigating how the algorithm reflects recent Hb changes and validating it with multiple measurements over time will be an important step toward determining clinical use cases of the algorithm.

To understand the underlying mechanisms of the model, we examined how the performance was affected by applying image ablation. First, we hypothesized that if the algorithm was based on the degree of pallor of the fundus as a whole, applying a Gaussian blur would have little effect on the performance. However, applying a Gaussian blur decreased the model performance, indicating that the model was dependent on the fine spatial features of the fundus images. In addition, we applied various masking methods to examine which portions of the fundus images are relevant to the model performance. Masking the top and bottom parts had little to no effect on the model performance, showing that the information contained in those areas was redundant. Masking the central core, which includes macula, had less effect than masking the horizontal stripe through the middle part, which includes both macula and disc. These results suggest that fine spatial features around the optic disc are crucial. Future studies could investigate this hypothesis further using multiple explanation techniques.

Another aspect of understanding the algorithm is whether it is detecting anemia or the underlying pathophysiology specific to each subtype of anemia. Anemia has multiple subtypes, and each has a different underlying etiology and requires different management. Other components of the CBC, in particular the MCV (average volume of red blood cells), are used to differentiate between the subtypes. For example, while iron-deficiency anemia typically presents with normal MCV, vitamin B12 or folate deficiency typically presents with elevated MCV, and anemia of chronic disease and thalassemia presents with decreased MCV. Thus, we hypothesized that if fundus images contained information about subtype-specific pathophysiology, the algorithm would be able to predict CBC results beyond Hb, RBC and HCT. However, the results did not support this hypothesis, indicating that the algorithm may be responding to features associated with the lack of hemoglobin itself. These results also illustrate that the algorithms may be useful for screening, but not for diagnosis. Patients would require referral and follow-up examinations such as blood tests before treatment.

When developing machine-learning-based algorithms, it is crucial to include a broad range of examples in the training set so that the developed algorithm generalize well in various settings. One of the limitations of our study is that we have used a dataset from a single source. For example, ethnicity is highly biased towards Caucasians in the UK Biobank dataset. Training and validating on multiple diverse datasets will be important for creating a generalizable algorithm.

To conclude, we showed that anemia and Hb level can be predicted from fundus images. Further research is warranted to examine if the approach is useful for scalable screening of anemia.

## Methods

### Study participants

The dataset for this study consisted of fundus images obtained from the UK Biobank[24], an observational study that recruited 500,000 participants, aged 40-69, across the United Kingdom between 2006 and 2010. The study was reviewed and approved by the North West Multi-Centre Research Ethics Committee. Each participant was consented and went through a series of health measurements and questionnaires. Each participant also provided blood, urine, and saliva samples. Approximately 70,000 patients also subsequently underwent ophthalmological examinations with paired retinal fundus and optical coherence tomography (OCT) imaging using a Topcon 3D OCT 1000 Mk2 (Topcon Corporation, Tokyo, Japan). Only retinal fundus images were included in this study. About 12% of the patients were excluded due to poor image quality as described previously[22]. Only patients with at least one fundus image paired with Hb measurement were included in this study (n=57,163). For complete blood count (CBC) analysis, only the patients who had all the CBC components measured were included (n=53,473). If a patient had multiple visits with paired fundus images and Hb measurement, only the first visit was included, except in the multiple-visit analysis described. The study was reviewed and approved by the North West Multi-Centre Research Ethics Committee. We randomly divided this dataset into a development set to develop our models (80%) and a validation set to assess our model's performance (20%) after stratifying patients by their gender and age. The validation set was not accessed during model development. 10% of the data in the development dataset ("tuning dataset") was used during model development for tuning hyper parameters such as learning rate and criterion for early stopping, with the remaining 70% ("training dataset") used for training the parameters of the neural networks.

### Definitions of anemia

Using WHO's guidelines[43], we used three sets of cut-offs based on Hb measurements: 12 g/dL for women and 13 g/dL for men (anemia), and 11 g/dL (moderate anemia). In addition, we assessed our results using a previously described gender neutral average anemia cutoff at 12.5 g/dL[9] for both men and women, which we call "approximate anemia".

### Categories of predictive models

In this study, we made two different types of predictions: continuous values (e.g. Hb or hematocrit; henceforth "regression tasks"), categorical values (e.g. presence or absence of anemia; henceforth “classification tasks”). Though a single model can in principle be trained for both regression tasks and multiple classification tasks, separate models were trained for regression tasks and classification tasks to keep the loss functions on a consistent scale. For each of these tasks, we compared the ability of three different categories of prediction models, each with a different set of input data. As a baseline, we used linear regression for regression tasks and logistic regression for classification tasks. These linear and logistic regression models used only demographic and clinical information (“metadata”, which are ethnicity, age, sex, current smoking status, systolic and diastolic blood pressure, pulse rate, height, weight, and body mass index (BMI)). We will refer to these as “metadata models”. Our second type of models used a deep convolutional neural network (details in the next section) with fundus images as input (“fundus-only models”). Our third and last type of models used both metadata and fundus images as input; the metadata was concatenated with the output of the Inception-v4 architecture[44] before the fully connected layer (“combined models”). Specifically, the fundus images were used as an input to a deep convolutional neural network (same structure as the fundus-only model), and the output of the convolutional neural network and metadata were provided to the combined model before the final layer (i.e., a “late fusion” model). The fully connected layer of fundus-only models and combined models had one output for each regression task and multiple outputs for each classification task (each output corresponds to each class).

## Development of the deep learning algorithms

Fundus images were preprocessed as previously described[23], while the input metadata and continuous output values (e.g. Hb) were standardized to have zero mean and unit variance. Using these data, a deep convolutional neural network with the Inception-v4 architecture[44] was developed and trained in TensorFlow[45]. The Inception-v4 network was initialized using parameters from a network pre-trained to classify objects in the ImageNet dataset[46], and the weights on the auxiliary connections from metadata were randomly initialized. Mean squared error was used as a loss function for regression tasks, and cross entropy was used for classification tasks. The models were trained with mini-batch stochastic gradient descent with momentum[47] with linear warm up[48] using Google Tensor Processing Unit (TPU) accelerators[49] (see supplementary methods). The learning rate was chosen to minimize the error in the tuning dataset. Since our network had a large number of parameters (43 million), to prevent overfitting training was terminated before convergence using early stopping[50] based on the performance on the tuning dataset. An ensemble of 10 networks[51] was trained on the same development set, and the outputs were averaged to yield the final prediction. For each patient, the final prediction was the average across both eyes.

### Evaluating the algorithms

To evaluate the model performance for continuous predictions, we used the mean absolute error (MAE) and $R^2$. For binary classification, we used the area under receiver operating characteristic curve (AUC) and sensitivity at various levels of specificity. To obtain 95% confidence intervals for these performance metrics, we used the non-parametric bootstrap procedure with 2,000 samples and reported the 2.5 and 97.5 percentiles.

### Ablation analysis

While applying ablation during both training and validation, we trained the fundus-only model for classification tasks and assessed the performance (AUC for predicting anemia and moderate anemia) of the model without ensembling or averaging across eyes. For each ablation method (e.g. masking 20% of the fundus at the center), 3 networks were trained, and the performance metrics were averaged across 3 networks.

## Code availability

The machine-learning models were developed using standard model libraries and scripts in TensorFlow[45]. Custom code was specific to our computing infrastructure and mainly used for data input/output and parallelization across computers.

## Data availability

The data that support the findings of this study are available, with restrictions, from UK Biobank[24].

## Acknowledgements

This research has been conducted using the UK Biobank Resource under Application Number 17643. We thank Dr. Christof Angermueller from Google Research for his engineering contribution, and Drs. Ali Zaidi, Arunachalam Narayanaswamy, Cameron Chen, Jonathan Krause, and Rory Sayres from Google Research for their advice and assistance with reviewing the manuscript.

## Author contributions

AM, GSC, LP, DRW, NH, and AVV designed the research. AM, LP, and AVV acquired data from UK Biobank. AM executed research and analyzed the data. AM, YL, and AVV interpreted the results. AM, AH, YL, and NH prepared the manuscript. All authors contributed to manuscript revision and approved the submitted version.

## Competing interests

The authors are employees of Google.

# Tables and figures

Table 1. Basic characteristics of the development datasets and the validation dataset.

| | Development Datasets | | Validation Dataset |
|---|---|---|---|
| | **Training dataset** | **Tuning dataset** | |
| Total no. images | 80,006 | 11,457 | 22,742 |
| No. of patients | 40,041 | 5,734 | 11,388 |
| Age (years)* | 57.9 (50.0-63.7) | 58.0 (49.9-63.7) | 57.9 (50.0-63.8) |
| Females (%) | 21,944 (54.8%) | 3,152 (55.0%) | 6,255 (54.9%) |
| Ethnicity (%) | | | |
| Black | 468 (1.2%) | 74 (1.3%) | 145 (1.3%) |
| Asian | 1,330 (3.3%) | 192 (3.3%) | 404 (3.5%) |
| White | 36,606 (91.4%) | 5,247 (91.5%) | 10,369 (91.1%) |
| Other | 1,637 (4.1%) | 221 (3.9%) | 470 (4.1%) |
| Current Smoker (%) | 3,794 (9.5%) | 544 (9.5%) | 1,120 (9.8%) |
| Body mass index (kg/$m^2$)* | 26.6 (24.0-29.7) | 26.7 (24.1-29.9) | 26.6 (24.1-29.7) |
| Height (cm)* | 168 (162-176) | 168 (162-176) | 168 (162-175) |
| Weight (kg)* | 76.3 (66.2-87.5) | 76.7 (66.4-87.6) | 76.3 (66.4-87.8) |
| Heart rate (bpm)* | 67.5 (61.0-74.5) | 67.5 (60.5-75.0) | 67.0 (61.0-74.0) |
| Systolic blood pressure (mmHg)* | 135 (124-148) | 135 (124-148) | 136 (124-148) |
| Diastolic blood pressure (mmHg)* | 82 (75-88) | 82 (75-89) | 82 (75-88) |
| Hemoglobin concentration (g/dL)* | 14.3 (13.4-15.2) | 14.3 (13.4-15.1) | 14.2 (13.4-15.1) |
| **Distribution of Anemia levels** | | | |
| None | 38,628 (96.5%) | 5,539 (96.6%) | 10,949 (96.1%) |
| Mild | 1,134 (2.8%) | 164 (2.9%) | 347 (3.0%) |
| Moderate | 267 (0.7%) | 31 (0.5%) | 90 (0.8%) |
| Severe | 12 (0.0%) | 0 (0.0%) | 2 (0.0%) |

* results are presented as median (interquartile range). bpm, beats per minute.

Table 2. Sensitivity at various levels of specificity.

| | Specificity | | |
|---|---|---|---|
| | 70% | 80% | 90% |
| Anemia | | | |
| Metadata-only | 63.3% (60.2-66.3) | 52.5% (48.6-56.1) | 32.0% (23.9-36.4) |
| Fundus-only | 86.0% (83.9-88.4) | 78.5% (74.5-82.4) | 64.0% (56.6-69.2) |
| Combined | **86.6%** (83.8-88.4) | **79.4%** (75.3-82.9)) | **65.2%** (60.5-71.1) |
| Moderate anemia | | | |
| Metadata-only | 78.0% (66.0-80.7) | 57.3% (47.6-70.5) | 39.7% (25.0-47.8) |
| Fundus-only | 94.8% (92.8-96.8) | 91.2% (87.7-94.4) | **86.9%** (80.8-90.8) |
| Combined | **96.7%** (93.0-97.8) | **92.3%** (88.1-95.9) | 84.2% (71.7-91.3) |
| Approximate anemia | | | |
| Metadata-only | 73.5% (72.2-75.1) | 65.8% (62.9-67.9) | 52.9% (50.8-55.2) |
| Fundus-only | 84.8% (82.6-86.9) | 76.4% (73.1-79.1) | 63.5% (59.2-67.5) |
| Combined | **86.1%** (85.0-88.0) | **80.9%** (78.6-83.1) | **69.7%** (66.9-73.6) |

Sensitivity is presented with 95% confidence intervals.

Bold indicates the highest sensitivity among the models at each condition.

Table 3. Accuracy of different methods to measure hemoglobin concentration.

| | Standard deviation of difference | 95% Limits of agreement | (Estimated) MAE (*) |
|---|---|---|---|
| Hematology analyzer[28] | 0.18 g/dL | 0.37 g/dL | 0.14 g/dL (*) |
| Invasive point-of-care devices[29] | 0.64 g/dL | NA | 0.51 g/dL (*) |
| Non-invasive point-of-care devices[29,30] | 1.4 g/dL | NA | 1.1 g/dL (*) |
| Smartphone-based method[9] | NA | 2.4 g/dL | 0.96 g/dL (*) |
| Our method (combined model) | 0.82 g/dL (0.80-0.85) | 1.58 g/dL (1.54-1.64) | 0.63 g/dL (0.62-0.64) |

(*) Mean absolute error (MAE) was estimated assuming that the error has a Gaussian distribution. Hematology analyzer was compared with hemoglobincyanide method. Other methods are compared with hematology analyzer. For the combined model, the results are shown with 95% confidence intervals.

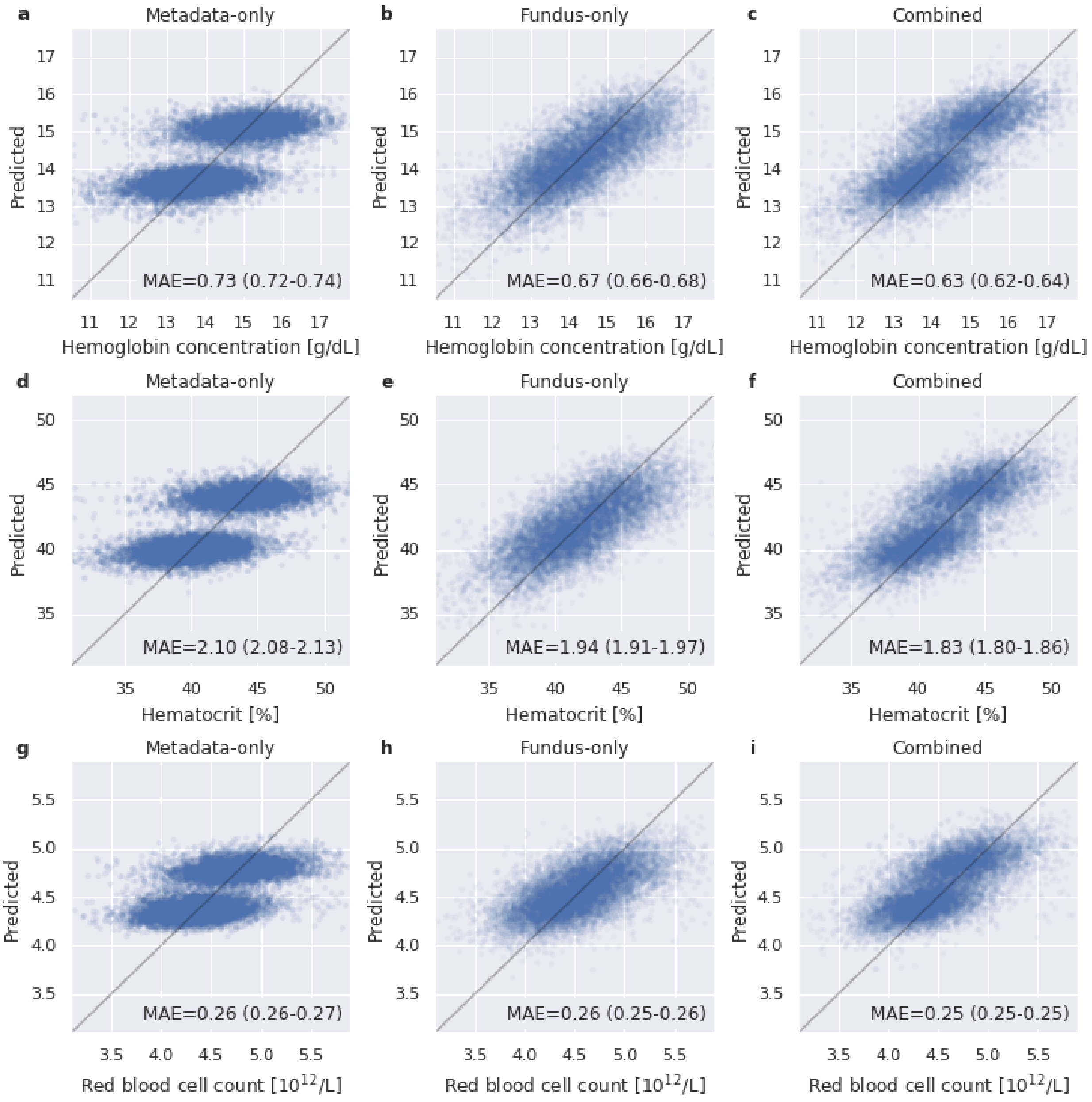

Figure 1. Prediction of hemoglobin concentration, hematocrit and red blood cell count.

**a**, Each blue dot represents each patient's measured hemoglobin concentration and predicted value from the metadata-only model. Gray line represents an identity line, where measured values equal predicted values. Inset text shows mean absolute error (MAE) with 95% confidence intervals (n=11,388). **b**, same as **a** for the fundus-only model. **c**, same as **a** for the combined model. **d**-**f**, same as **a**-**c** for hematocrit. **g**-**i**, same as **a**-**c** for red blood cell count.

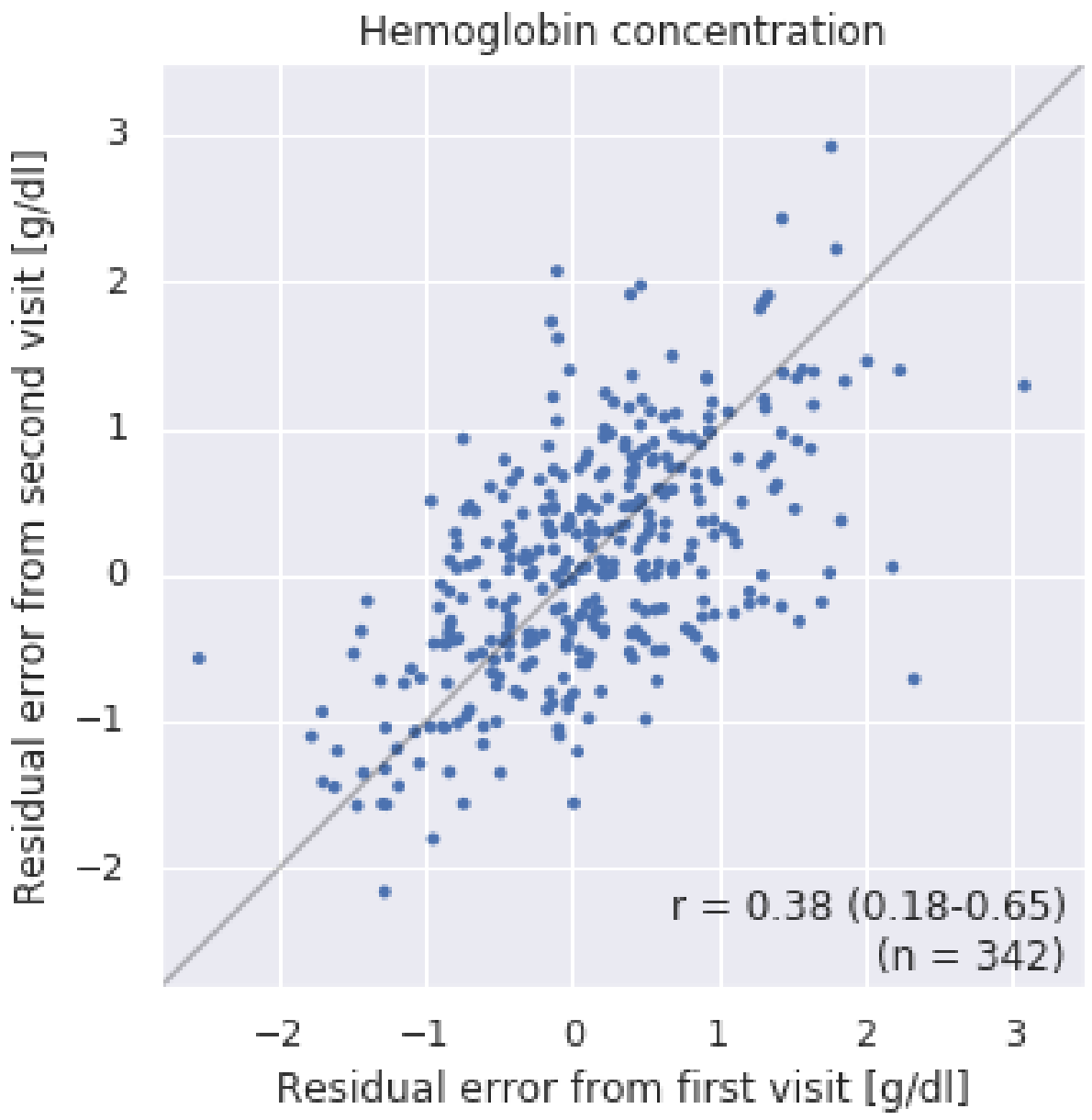

Figure 2. Correlation between residual errors of the combined model from multiple visits.

Each blue dot represents each patient (n=342). Residual error is the difference between measured hemoglobin concentration and prediction by the combined model. Gray line represents an identity line. Pearson's correlation coefficient r = 0.38 (95% confidence interval: 0.18-0.65).

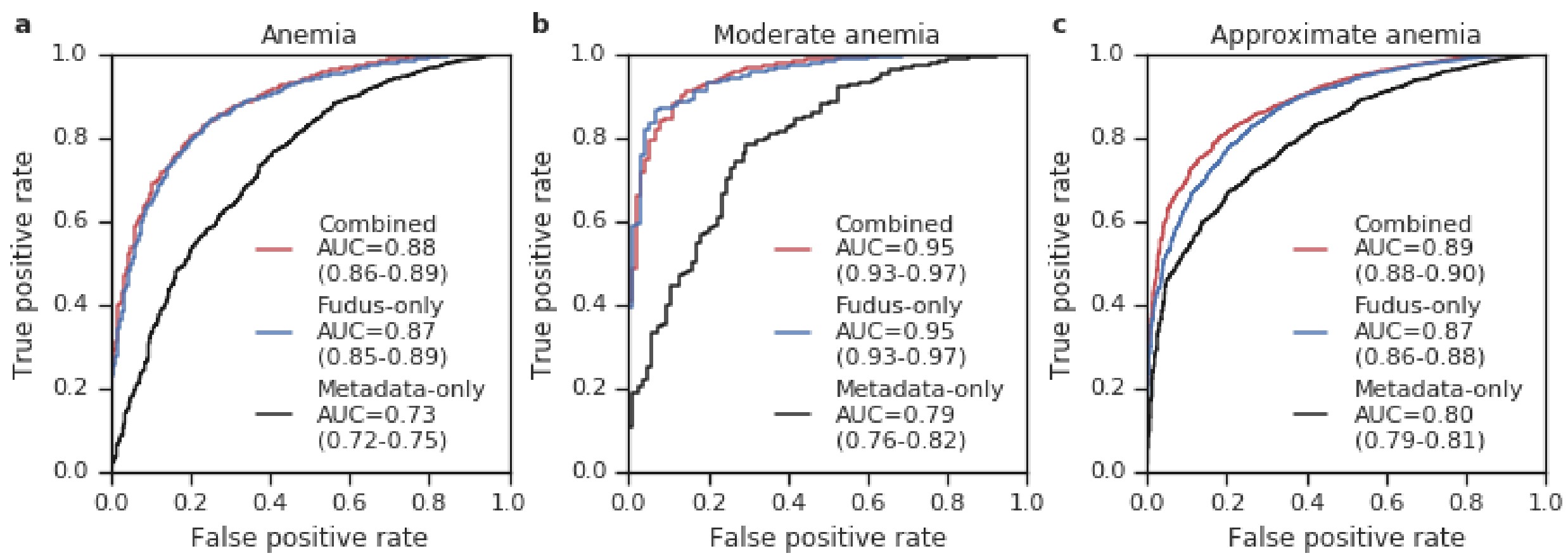

Figure 3. Prediction of anemia classifications.

**a**, Receiver operating characteristic (ROC) curves for detecting anemia by the metadata-only model (black line), the fundus-only model (blue line), and the combined model (red line). Area under the curve (AUC) for each model is shown with 95% confidence intervals (n=11,388). **b**, ROC curves and AUCs for detecting moderate anemia. **c**, ROC curves and AUCs for detecting approximate anemia (Methods).

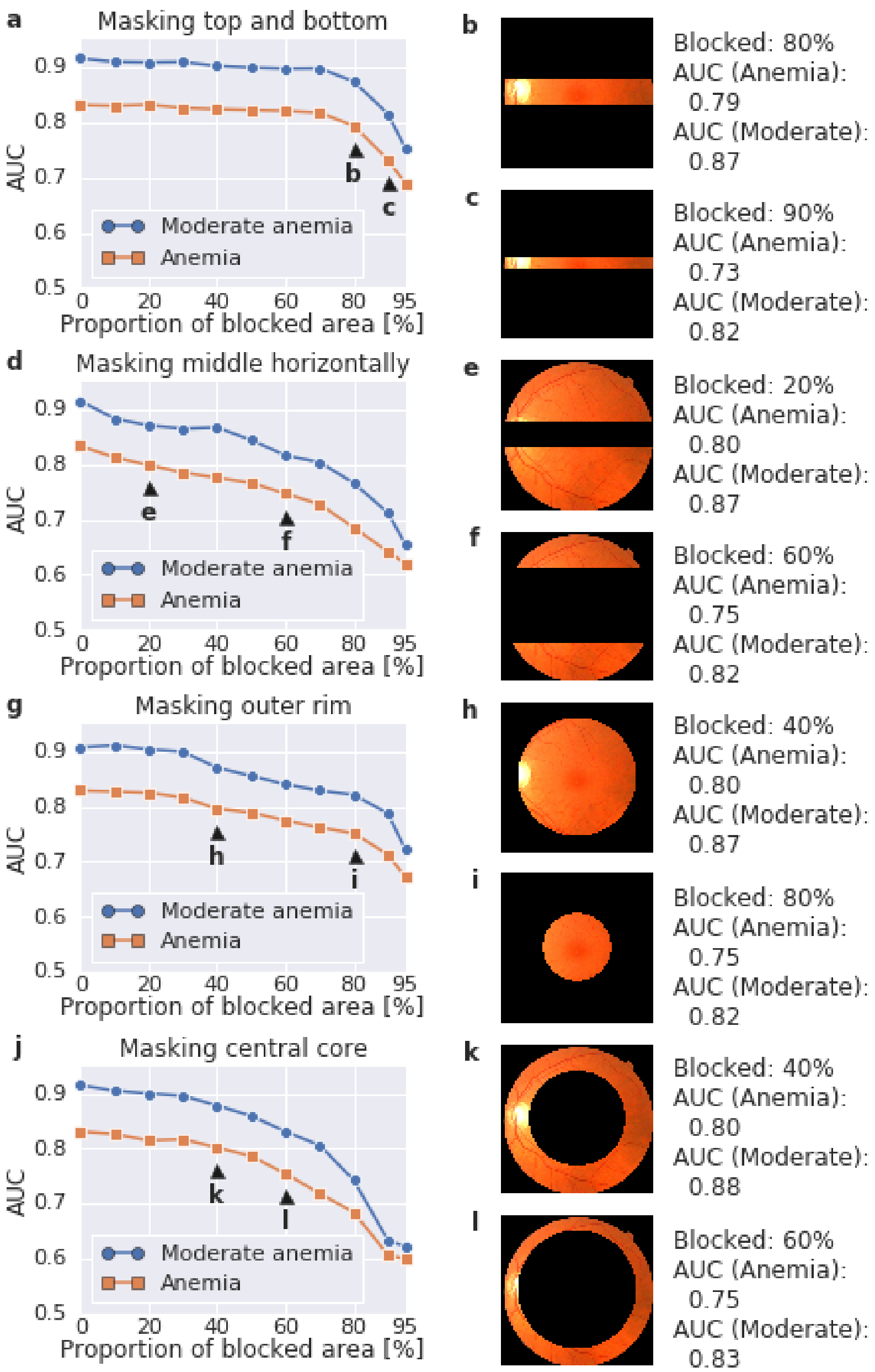

Figure 4. Effects of masking parts of the image on the prediction of anemia and moderate anemia.

The masking was applied during both training and validation. **a**, Masking the top and bottom parts of the images. Arrowheads correspond to examples shown in **b** and **c**. **b** and **c**, Example masked images. **d-f**, Similar to **a-c**, but for masking a horizontal stripe through the middle of the images. **g-i**, Similar to **a-c**, but for masking the outer rim of the images. **j-l**, Similar to **a-c**, but for masking a central core of the images. AUC, area under receiver operating characteristic curve.

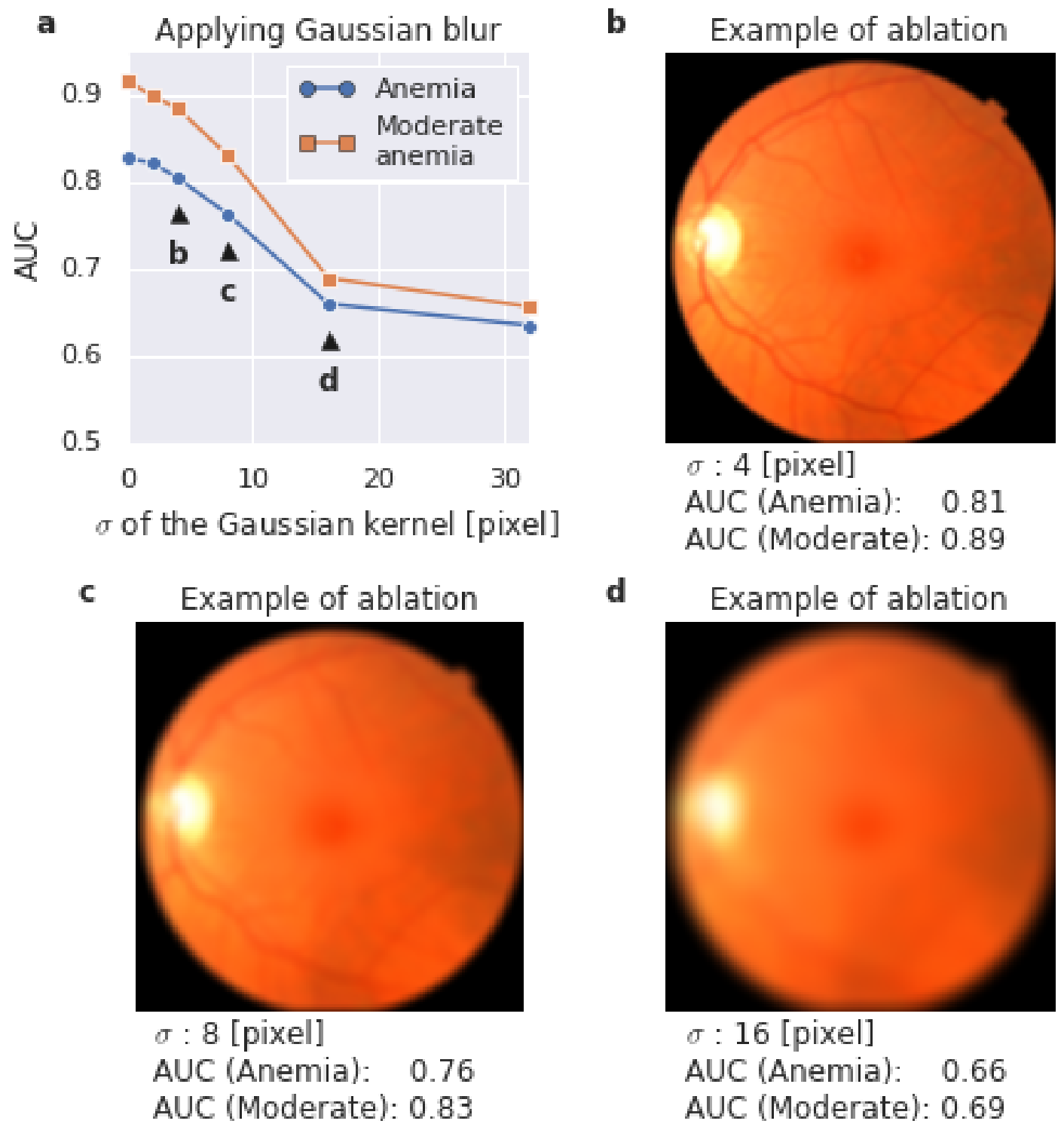

Figure 5. Effects of removing high-frequency information using Gaussian blur on the prediction of anemia and moderate anemia.

The ablation was applied during both training and validation. **a**, Model performance as a function of a Gaussian blur amount. Arrowheads correspond to examples shown in **b**-**d**. **b**-**d**, Example images with varying blur amounts. σ, standard deviation of the Gaussian kernel used for a blur, which was applied after images were resized to 587 pixel × 587 pixel. AUC, area under receiver operating characteristic curve.

# Supplementary information

## Supplementary methods

We trained each model on a single Google TPUv2 unit (8 cores across 4 chips, 8 GB memory per core, 180 teraFLOPS in total) with the following hyperparameters:
– Inception-v4 architecture
– Input image resolution: 587 × 587
– Learning rate: 0.0008
    – Learning rate was initially 0.0001, and linearly increased at each step until it reached 0.0008 after 3 epochs[48].
– Batch size: 64 (8 per core, and 8 cores ran synchronously)
– Batch size for batch normalization: 8 (batch normalization was performed at each core)
– Weight decay: 0.00004
– Mini-batch stochastic gradient descent optimizer with momentum[47]
– Data augmentation (in order):
  • Random vertical and horizontal reflections
  • Random brightness changes with a max delta of 0.1147528
  • Random saturation changes between 0.5597273 and 1.2748845
  • Random hue changes between -0.0251488 and 0.0251488
  • Random contrast changes between 0.9996807 and 1.7704824
– For classification tasks, each class was weighted in proportion to the number of examples of the class to the power of -0.9.
– Model evaluations were performed using a running average of parameters, with an exponential decay factor of 0.9999.

## Table S1. Prediction of complete blood count components.

The metadata-only model, the fundus-only model, and the combined model were trained to predict all the complete blood components. Note that the difference from Figure 1 is due to being trained for additional components at the same time and using a smaller subset of patients who had all the components measured. $R^2$, coefficient of determination.

| Complete blood count components | $R^2$ | | | Mean absolute error | | | |
|---|---|---|---|---|---|---|---|
| | Metadata-only | Fundus-only | Combined | Metadata-only | Fundus-only | Combined | unit |
| Hemooglobin concentration | 0.43 (0.41-0.44) | 0.51 (0.50-0.52) | 0.52 (0.51-0.53) | 0.95 (0.93-0.96) | 0.69 (0.68-0.70) | 0.67 (0.66-0.68) | g/dL |
| Hemoatcrit percentage | 0.41 (0.40-0.42) | 0.47 (0.46-0.48) | 0.49 (0.48-0.50) | 2.70 (2.66-2.74) | 2.02 (1.99-2.05) | 1.96 (1.93-1.99) | % |
| Red blood cell count | 0.32 (0.31-0.33) | 0.35 (0.34-0.36) | 0.36 (0.34-0.37) | 0.34 (0.33-0.34) | 0.26 (0.25-0.26) | 0.26 (0.25-0.26) | $10^{12}$/L |
| High light scatter reticulocyte count | 0.18 (0.16-0.19) | 0.05 (0.04-0.06) | 0.18 (0.17-0.20) | 0.01 (0.01-0.01) | 0.01 (0.01-0.01) | 0.01 (0.01-0.01) | $10^{12}$/L |
| Reticulocyte count | 0.16 (0.15-0.17) | 0.05 (0.04-0.06) | 0.17 (0.15-0.18) | 0.02 (0.02-0.02) | 0.02 (0.02-0.02) | 0.02 (0.02-0.02) | $10^{12}$/L |
| High light scatter reticulocyte percentage | 0.14 (0.13-0.16) | 0.03 (0.02-0.04) | 0.15 (0.13-0.16) | 0.20 (0.19-0.21) | 0.15 (0.15-0.16) | 0.14 (0.14-0.15) | % |
| Mean corpuscular hemooglobin | 0.11 (0.09-0.12) | 0.09 (0.08-0.11) | 0.12 (0.11-0.14) | 1.79 (1.73-1.85) | 1.29 (1.26-1.31) | 1.27 (1.24-1.29) | pg |
| Neutrophill count | 0.11 (0.10-0.12) | 0.04 (0.03-0.05) | 0.12 (0.11-0.13) | 1.33 (1.30-1.36) | 1.08 (1.06-1.10) | 1.00 (0.99-1.02) | $10^9$/L |
| Immature reticulocyte fraction | 0.12 (0.11-0.13) | 0.04 (0.03-0.04) | 0.12 (0.11-0.13) | 0.06 (0.06-0.06) | 0.05 (0.05-0.05) | 0.04 (0.04-0.05) | ratio |
| Reticulocyte percentage | 0.12 (0.10-0.13) | 0.01 (0.00-0.02) | 0.12 (0.10-0.13) | 0.49 (0.48-0.51) | 0.39 (0.38-0.40) | 0.37 (0.36-0.38) | % |
| Mean corpuscular volume | 0.11 (0.09-0.12) | 0.07 (0.06-0.08) | 0.12 (0.10-0.13) | 4.35 (4.24-4.47) | 3.26 (3.20-3.32) | 3.17 (3.11-3.23) | fL |
| Platelet crit | 0.11 (0.10-0.12) | 0.07 (0.06-0.08) | 0.12 (0.10-0.13) | 0.04 (0.04-0.04) | 0.03 (0.03-0.03) | 0.03 (0.03-0.03) | % |
| Lymphocyte percentage | 0.09 (0.08-0.10) | 0.04 (0.02-0.05) | 0.09 (0.08-0.10) | 7.27 (7.14-7.41) | 5.81 (5.72-5.90) | 5.64 (5.55-5.73) | % |
| Platelet count | 0.09 (0.08-0.10) | 0.04 (0.03-0.05) | 0.09 (0.08-0.10) | 52.89 (51.86-53.94) | 41.34 (40.69-42.00) | 40.19 (39.58-40.83) | $10^9$/L |
| White blood cell count | 0.08 (0.04-0.12) | 0.03 (0.02-0.05) | 0.08 (0.04-0.13) | 2.16 (1.71-2.78) | 1.41 (1.38-1.44) | 1.31 (1.28-1.35) | $10^9$/L |
| Neutrophill percentage | 0.07 (0.06-0.08) | 0.02 (0.01-0.03) | 0.07 (0.06-0.08) | 8.24 (8.09-8.39) | 6.56 (6.46-6.66) | 6.38 (6.28-6.47) | % |
| Red blood cell distribution width | 0.03 (0.02-0.04) | 0.08 (0.07-0.10) | 0.07 (0.06-0.08) | 0.99 (0.95-1.04) | 0.65 (0.64-0.66) | 0.64 (0.63-0.65) | % |
| Mean sphered cell volume | 0.07 (0.06-0.07) | -0.00 (-0.01-0.01) | 0.07 (0.06-0.08) | 5.10 (4.99-5.21) | 4.12 (4.06-4.18) | 3.91 (3.84-3.97) | fL |
| Mean corpuscular hemoglobin concentration | 0.05 (0.04-0.06) | 0.06 (0.05-0.07) | 0.06 (0.05-0.07) | 0.79 (0.74-0.85) | 0.56 (0.55-0.57) | 0.56 (0.55-0.57) | g/dL |
| Monocyte count | 0.06 (0.04-0.07) | 0.03 (0.02-0.04) | 0.06 (0.04-0.08) | 0.20 (0.18-0.22) | 0.13 (0.13-0.13) | 0.12 (0.12-0.13) | $10^9$/L |
| Monocyte percentage | 0.04 (0.03-0.05) | 0.02 (0.01-0.02) | 0.04 (0.03-0.06) | 2.60 (2.30-2.93) | 1.63 (1.60-1.68) | 1.59 (1.56-1.63) | % |
| Platelet distribution width | 0.03 (0.02-0.04) | 0.02 (0.02-0.03) | 0.03 (0.03-0.04) | 0.51 (0.50-0.52) | 0.40 (0.39-0.41) | 0.40 (0.39-0.40) | % |
| Nucleated red blood cell percentage | 0.02 (0.01-0.03) | -0.01 (-0.02-0.00) | 0.02 (0.01-0.03) | 0.34 (0.27-0.41) | 0.05 (0.04-0.06) | 0.06 (0.05-0.07) | % |
| Mean reticulocyte volume | 0.02 (0.01-0.03) | 0.00 (-0.01-0.01) | 0.02 (0.02-0.03) | 8.05 (7.87-8.25) | 6.21 (6.11-6.32) | 6.12 (6.02-6.22) | fL |
| Nucleated red blood cell count | 0.02 (0.01-0.03) | -0.02 (-0.03--0.01) | 0.02 (0.01-0.03) | 0.02 (0.02-0.03) | 0.00 (0.00-0.00) | 0.00 (0.00-0.00) | $10^9$/L |
| Eosinophill count | 0.02 (0.01-0.03) | 0.01 (0.01-0.02) | 0.02 (0.01-0.03) | 0.13 (0.12-0.14) | 0.09 (0.09-0.09) | 0.09 (0.08-0.09) | $10^9$/L |
| Eosinophill percentage | 0.01 (0.01-0.02) | 0.00 (0.00-0.01) | 0.01 (0.01-0.02) | 1.72 (1.65-1.79) | 1.22 (1.20-1.25) | 1.20 (1.18-1.23) | % |
| Lymphocyte count | 0.02 (0.00-0.07) | 0.00 (-0.00-0.02) | 0.01 (0.00-0.07) | 1.44 (0.72-2.22) | 0.52 (0.49-0.55) | 0.50 (0.48-0.53) | $10^9$/L |
| Mean platelet volume | 0.01 (0.01-0.02) | 0.01 (0.00-0.01) | 0.01 (0.01-0.02) | 1.08 (1.06-1.10) | 0.85 (0.84-0.87) | 0.85 (0.83-0.86) | fL |
| Basophill count | 0.01 (0.00-0.01) | 0.00 (0.00-0.01) | 0.01 (0.01-0.01) | 0.05 (0.05-0.06) | 0.03 (0.02-0.03) | 0.02 (0.02-0.03) | $10^9$/L |
| Basophill percentage | 0.00 (0.00-0.01) | 0.00 (0.00-0.00) | 0.00 (0.00-0.01) | 0.70 (0.61-0.79) | 0.34 (0.32-0.35) | 0.33 (0.32-0.34) | % |